%% file: main.tex
\documentclass[11pt,letterpaper]{article}

\usepackage[margin=0.75in]{geometry}
\usepackage{newtxtext}
\usepackage{amsmath,amssymb}
\usepackage[hyphens]{url}
\usepackage{graphicx}
\usepackage[numbers,sort&compress]{natbib}
\usepackage{caption}
\usepackage{booktabs}
\usepackage{multirow}
\usepackage{array}
\usepackage{longtable}
\usepackage{pifont}
\usepackage{arydshln}
\usepackage{placeins}
\usepackage[colorlinks=true,allcolors=blue]{hyperref}

\hypersetup{
  pdftitle={PuzzleKV: Page-Wise Low-Rank Decomposition for KV Cache Compression},
  pdfauthor={Zizhong Wang, Jieying Wang, Zhao Zhang, Jiajia Li}
}

\newcommand{\methodname}{PuzzleKV}

\title{PuzzleKV: Page-Wise Low-Rank Decomposition for KV Cache Compression}

\author{Zizhong Wang, Jieying Wang, Zhao Zhang, Jiajia Li}
\date{}

\begin{document}

\maketitle

\begin{abstract}

Long-context inference in large language models (LLMs) is increasingly
limited by the memory required for the key-value (KV) cache. KV cache
compression addresses this problem by reducing the storage cost of
previous tokens. Among existing approaches, low-rank compression is
particularly attractive because it represents every token in reduced
dimensions. Previous low-rank methods typically derive fixed projection spaces
from model weights, construct fixed spaces from calibration activations, or
construct a shared basis over a broad cache region. Such representations may not capture detailed but important information. We partition each per-head KV cache into
fixed-length logical pages and observe substantial low-rank structure
within individual pages. 
Based on this observation, we propose \methodname{},
a training- and calibration-free method that treats each completed page
as an independent compression unit. \methodname{} decomposes pages within
each layer and KV head, computes attention directly over dense and
factorized pages, and incrementally compresses newly eligible pages
during autoregressive decoding.
Experiments across models, context lengths, and benchmarks demonstrate
the effectiveness of \methodname{} under matched storage budgets. At approximately 60\%
of the original KV cache storage, \methodname{} achieves more than 96\% of
Full KV performance across both evaluated models and all benchmark
settings, with substantial gains over Global SVD on RULER and
competitive performance on LongBench. To achieve a more aggressive compression ratio,
\methodname{} can be further combined with
quantization while retaining more than 93\% of Full KV performance using
only 18.7\% of the original storage.
\end{abstract}

\section{Introduction}
Large language models increasingly rely on few-shot demonstrations,
chains of thought, and interleaved reasoning and action to solve
complex tasks~\citep{tom2020language,wei2022chain,yao2023react}. As
these inputs grow, longer context windows allow LLMs to access more
task-relevant information, but they also place substantial memory
pressure on inference.

Autoregressive decoders store the attention keys and values of previous
tokens in a key-value (KV) cache to avoid recomputing them at every
generation step~\citep{vaswani2017attention,kwon2023efficient}. For a
model with $L$ layers, $H_{kv}$ KV heads, head dimension $d$, and
context length $T$, a single request holds $2LH_{kv}Td$ KV elements
before batching. Qwen3-32B~\citep{yang2025qwen3}, for example, has 64 layers, 8 KV heads, and
head dimension 128, so under BF16 a single
128K-token sequence requires about 32~GiB of KV cache. Because this
footprint grows linearly with both context length and serving batch
size, KV cache compression is essential for efficient long-context
inference.

Existing methods reduce KV cache storage through token eviction,
quantization, or low-rank decomposition. Eviction discards selected
tokens~\citep{zhang2023h2o} and quantization stores keys and values at
lower precision~\citep{liu2024kivi}, whereas low-rank methods represent
keys and values in lower-dimensional spaces. The latter differ mainly
in the source of the compression space: some factorize model projection
weights~\citep{chang2024palu,zhang2024lorc}, often requiring calibration
or per-model offline processing, while others build or update a basis
from the KV cache produced at inference~\citep{zhu2026ojakv,chang2025xkv}.
In either case, a single basis is shared across a broad cache region,
favoring directions that dominate overall reconstruction and potentially
missing sparse but task-critical information. 

To capture this local structure at a finer granularity, we take inspiration from the fixed-size blocks used by PagedAttention to manage KV cache memory~\citep{kwon2023efficient} and reinterpret the logical page as a unit of compression. This choice is well founded: our analysis shows
substantial low-rank redundancy within individual key and value pages
across models, layers, and KV heads (Figure~\ref{fig:page-effective-rank}).
Using the page as the compression unit lets a single abstraction govern
both low-rank representation and incremental cache updates, while the
uniform shape and independence of pages allow their decompositions to be
processed in batches on GPUs.

Based on this design, we propose \methodname{}, a training- and calibration-free
page-wise low-rank KV cache compression method. \methodname{} independently decomposes completed pages within each layer and
KV head, while keeping a small sink region and recent window in dense
form. Its mixed attention kernel directly processes dense and
factorized pages without reconstructing the historical KV cache. During
autoregressive decoding, newly eligible pages are incrementally
converted into factor storage.

Our main contributions are as follows:
\begin{itemize}
    \item We introduce \methodname{}, a training- and calibration-free method
    that independently factorizes completed KV pages within each layer
    and KV head, preserving every token through page-local low-rank
    representations. 
    \item We implement \methodname{} with batched page
    decomposition, direct attention over dense and factorized pages
    without reconstruction, and incremental page conversion during
    decoding.
    \item We evaluate \methodname{} on Qwen3-8B and Llama-3.1-8B-Instruct across
    RULER and LongBench. At approximately 60\% of Full KV storage, \methodname{} achieves more than 96\% of
    Full KV performance across all evaluated settings, establishes
    substantial gains over Global SVD on RULER, and achieves competitive
    performance on LongBench. Combined with
    per-factor INT4 quantization, it uses only 18.7\% of the original
    storage while achieving more than 93\% of Full KV performance.

\end{itemize}

\section{Related Work}

KV cache compression is essential for memory-efficient LLM inference, and
existing methods broadly fall into three categories: low-rank decomposition, quantization,
and eviction~\citep{li2024survey}.

\paragraph{Low-Rank KV Cache Compression.}
Low-rank methods reduce KV cache storage by representing keys and values in lower-dimensional spaces. Palu~\citep{chang2024palu} and LoRC~\citep{zhang2024lorc} factorize the projection matrices, allowing the model to cache low-dimensional intermediate representations. ECKVH~\citep{yu2024eckvh} and EigenAttention~\citep{saxena2024eigenattention} construct fixed compression bases from activations collected on a calibration dataset. OjaKV~\citep{zhu2026ojakv} updates a sequence-level low-rank basis online, while xKV~\citep{chang2025xkv} exploits cross-layer redundancy during prefill. \methodname{} instead independently factorizes each completed page within each layer and KV head.

\paragraph{KV Cache Quantization.}
Quantization reduces KV cache memory by storing keys and values at lower numerical precision, using schemes such as asymmetric low-bit quantization~\citep{liu2024kivi}, polar transformations of angular components~\citep{han2025polarquant}, and distortion-aware vector quantization~\citep{zandieh2025turboquant}. 
\methodname{} instead reduces the dimensionality of each KV page via low-rank decomposition, and remains complementary to quantization, which can be applied to its low-rank factors.

\paragraph{KV Cache Eviction.}
Eviction methods reduce KV cache storage by keeping only tokens expected to remain useful, but an evicted token is permanently lost for future inference. H2O~\citep{zhang2023h2o} retains heavy-hitter and recent tokens, StreamingLLM~\citep{xiao2023streamingllm} keeps attention sinks with a local window, and SnapKV~\citep{li2024snapkv} selects important positions from prompt-time attention.
Unlike eviction methods, \methodname{} retains every token in compressed form.

\section{Methodology}
\label{sec:method}

\begin{figure}[t]
  \centering
  \includegraphics[width=0.84\textwidth]{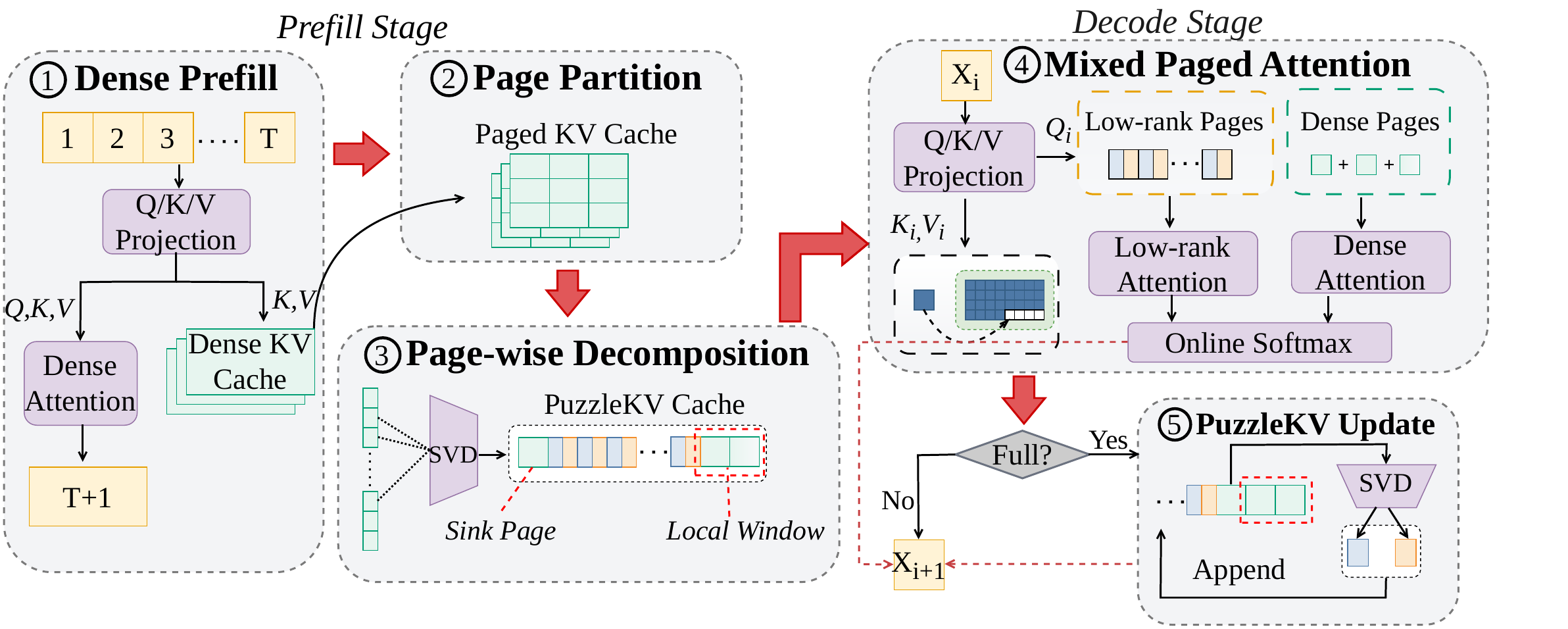}
  \caption{Overview of \methodname{}. The workflow covers page partitioning and
  page-wise decomposition during prefill, followed by mixed paged attention
  and incremental cache updates during decoding.}
  \label{fig:padekv-overview}
\end{figure}

\subsection{Design Principle: Page as a Local Subspace}
\begin{figure}[t]
  \centering
  \includegraphics[width=0.46\textwidth]{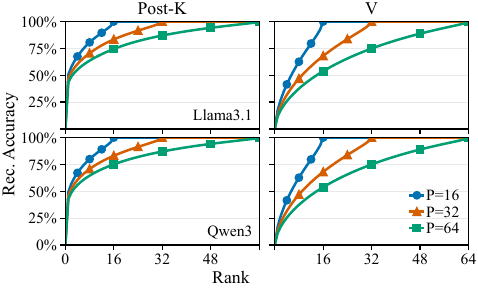}%
  \hspace{0.01\textwidth}%
  \includegraphics[width=0.46\textwidth]{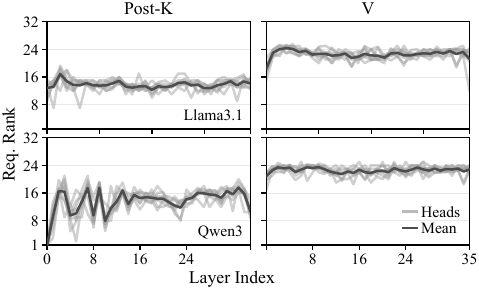}
  \caption{Page-level low-rank structure across models, layers, KV heads,
  and page sizes. Left: Reconstruction accuracy under different ranks
  and page sizes. Right: Required rank for 80\% reconstruction accuracy
  across layers at $P=32$, with individual KV heads shown in thin lines
  and their mean in bold.}
  \label{fig:page-effective-rank}
\end{figure}

A key design choice in low-rank KV compression is the granularity at which tokens share a low-rank basis. A single basis spanning the entire prompt maximizes reuse, while the sequence itself spans regions with distinct dominant directions. Assigning each KV page its own compact basis lets every region be represented by the directions most relevant to it, yielding fine-grained low-rank representations that together cover a richer set of directions across the sequence while keeping each page strongly low-rank. This design pays off when individual pages are themselves low-rank, which we confirm by examining how accurately KV pages reconstruct across different ranks and page sizes in Qwen3-8B and Llama-3.1-8B-Instruct. As shown in Figure~\ref{fig:page-effective-rank}, page-level low-rank structure is consistently observed across both models and all evaluated page sizes.
At $P=32$, although the required ranks vary across layers and KV heads, they remain below the full page rank across all evaluated layers and heads. Together, these results establish page-level low-rank structure as a consistent property across the evaluated models, page sizes, layers, and KV heads.

\subsection{Preliminaries: Low-Rank KV Cache}
\label{sec:preliminaries}

\paragraph{KV Cache and Attention.}
At each Transformer layer, the key and value projections produce
$K,V\in\mathbb{R}^{H_{kv}\times T\times d}$, where $H_{kv}$ is the
number of KV heads, $T$ is the sequence length, and $d$ is the head
dimension. During autoregressive decoding, each new key--value pair is
appended to the cache along the sequence dimension. 
Omitting the head index for clarity, attention for a query $Q$ is
computed as
{\small
\begin{equation}
\operatorname{Attn}(Q,K,V)
=
\operatorname{softmax}
\left(
\frac{QK^{\top}}{\sqrt{d}}
\right)V.
\label{eq:dense-attention}
\end{equation}
}

\paragraph{Truncated SVD.}
For a matrix $X\in\mathbb{R}^{n\times d}$ with singular value
decomposition $X=U\Sigma W^{\top}$, where the singular values in
$\Sigma$ are arranged in descending order, its rank-$r$ truncated
approximation retains the leading $r$ components,
$
\widehat{X}_r
=
U_r\Sigma_r W_r^{\top}.
$
Here, $U_r\in\mathbb{R}^{n\times r}$,
$\Sigma_r\in\mathbb{R}^{r\times r}$, and
$W_r\in\mathbb{R}^{d\times r}$. By the
Eckart--Young--Mirsky theorem, $\widehat{X}_r$ minimizes
$\|X-\widehat{X}\|_F$ over all matrices $\widehat{X}$ with rank at
most $r$.

\subsection{Method: \methodname{}}
Figure~\ref{fig:padekv-overview} summarizes the \methodname{} workflow, which spans a prefill stage (left) and a decoding stage (right) that operate on a single hybrid KV cache.

During prefill, \methodname{} \ding{172}~computes the standard dense KV cache, \ding{173}~partitions each layer and KV head into fixed-size pages, and \ding{174}~factorizes every completed page while keeping the attention sink and a local window in dense form. The result is a hybrid cache in which sink and recent pages stay dense while historical pages are stored as low-rank factors. During decoding, \ding{175}~each query attends jointly over dense and factorized pages and merges their partial results through online softmax without reconstructing the historical cache; then \ding{176}~the new key--value pair is appended to the most recent dense page, and once that page fills and leaves the local window it is factorized and moved into low-rank storage. Steps \ding{175} and \ding{176} form the steady-state decoding loop: the mixed-attention kernel reads the hybrid cache that the incremental update continually maintains, keeping every token in compressed form throughout generation.

\paragraph{Page-Wise Low-Rank KV Cache Construction \ding{172}, \ding{173}, \ding{174}.}
\methodname{} treats each completed KV page as an independent low-rank subspace while retaining every token in compressed form. During prefill, it computes the standard dense KV cache and partitions the tokens at each layer and KV head into fixed-size pages of $P$ tokens. \methodname{} then retains two regions in dense form: an initial \emph{sink page} holding the first tokens, which attract a disproportionate share of attention and are cheap to keep uncompressed~\citep{xiao2023streamingllm}, and a moving \emph{local window} of the most recent pages, whose tokens are accessed most frequently and whose current page is still being filled during decoding. Every other completed page is independently factorized with truncated SVD, replacing each eligible dense page by two factors:
{\small
\begin{equation}
\begin{aligned}
\widehat{X}_p
&=
L_p^{(X)}R_p^{(X)}, \\
L_p^{(X)}
&=
U_{p,r_X}^{(X)}, \qquad
R_p^{(X)}
=
\Sigma_{p,r_X}^{(X)}
\left(W_{p,r_X}^{(X)}\right)^{\top},
\end{aligned}
\label{eq:padekv-factor-format}
\end{equation}
}
where $X\in\{K,V\}$, with $r_X=r_K$ for keys and $r_X=r_V$ for values. The resulting cache holds dense sink and recent pages alongside factorized historical pages, discarding no token. Excluding pages stored in dense form, the storage ratio of a factorized key--value page pair is
{\small
\begin{equation}
\rho_{\mathrm{page}}
=
\frac{(r_K+r_V)(P+d)}{2Pd}.
\label{eq:page-storage-ratio}
\end{equation}
}

To obtain these factors efficiently, \methodname{} avoids invoking a separate SVD on every $P\times d$ page. Because KV pages are short and wide ($P<d$) and share the same shape, it batches them and recovers the leading left singular subspace from the smaller $P\times P$ Gram matrix $G_p=X_pX_p^{\top}$, taking $L_p^{(X)}=Q_{p,r_X}$ as its top-$r_X$ eigenvectors and $R_p^{(X)}=Q_{p,r_X}^{\top}X_p$. Since the eigenvectors of $X_pX_p^{\top}$ are the left singular vectors of $X_p$, this yields exactly the truncated-SVD factors above while mapping to batched GPU operations.

\paragraph{Attention over Mixed KV Pages \ding{175}.}
\methodname{} computes attention over dense and factorized pages through a custom mixed-attention path, evaluating each factorized page directly from its stored factors and never reconstructing the historical dense cache. This avoids the transient dense buffers that explicit reconstruction would create, which prior work identifies as a source of lost memory savings~\citep{shao2026flashsvd}. For a query $q$ and a factorized page $\widehat{K}_p=L_p^{(K)}R_p^{(K)}$, $\widehat{V}_p=L_p^{(V)}R_p^{(V)}$, \methodname{} computes the page-local softmax state
{\small
\begin{equation}
\begin{aligned}
&s_p=
\frac{
\left(q(R_p^{(K)})^\top\right)
(L_p^{(K)})^\top
}{\sqrt d}, \\
&m_p=\max_j s_{p,j}, \\
&\ell_p=\sum_j \exp(s_{p,j}-m_p), \\
&o_p=\left(\exp(s_p-m_p)L_p^{(V)}\right)R_p^{(V)}.
\end{aligned}
\label{eq:factorized-kv-attention}
\end{equation}
}
where $s_p$ are the attention scores, $m_p$ and $\ell_p$ the running maximum and normalizer, and $o_p$ the unnormalized page output. The parenthesization contracts $q$ with the small factors first, so the reconstructed $\widehat{K}_p$ and $\widehat{V}_p$ are never materialized. Dense sink and recent pages produce the same state $(m_p,\ell_p,o_p)$ from their stored $K_p$ and $V_p$. \methodname{} then merges all pages through the online-softmax reduction of FlashAttention~\citep{dao2022flashattention}:
{\small
\begin{equation}
\begin{aligned}
&m=\max_p m_p, \\
&o=
\frac{
\sum_p \exp(m_p-m)o_p
}{
\sum_p \exp(m_p-m)\ell_p
}.
\end{aligned}
\label{eq:mixed-page-reduction}
\end{equation}
}
This yields globally normalized attention over dense and factorized pages without materializing any factorized historical page.

\paragraph{Incremental Cache Update \ding{176}.}
\methodname{} extends compression from prefill to the entire decoding process, keeping the hybrid cache current as generation proceeds rather than compressing only once. Each new key--value pair is appended to the most recent dense page and immediately participates in attention. When a decoding step fills that page, it enters the dense local window, and the oldest page leaving the window is factorized once, with its factors written to preallocated storage and its dense slot released. Previously factorized pages are never revisited, so each step incurs at most one page factorization and every generated token remains in compressed form.

\section{Experiments}
\label{sec:experiments}

We evaluate \methodname{} across two models and long-context benchmarks along
three axes: quality preservation under a matched storage budget,
sensitivity to its key design choices, and the runtime overhead and
logical KV cache footprint of our implementation. We first present the
experimental setup and main accuracy results, followed by quantization
compatibility, ablation studies, and runtime and memory measurements.

\subsection{Experimental Setup}
\label{sec:experimental-setup}

\paragraph{Models and Benchmarks.}
We evaluate \methodname{} on Qwen3-8B (Qwen3;~\citealp{yang2025qwen3}) and
Llama-3.1-8B-Instruct (Llama3.1;~\citealp{grattafiori2024llama}) across two
complementary long-context benchmarks. RULER~\citep{hsieh2024ruler},
evaluated at input lengths of 16K and 32K, measures
compression-induced degradation on controlled synthetic tasks, while
LongBench~\citep{bai2024longbench} assesses performance on real-world
long-context tasks after compression.

\paragraph{Baselines.}
We compare against two representative compression methods and one
controlled baseline. Palu~\citep{chang2024palu} represents
calibration-based, weight-side low-rank compression, and
H2O~\citep{zhang2023h2o} represents token eviction. Global SVD is our
controlled baseline for isolating the effect of compression
granularity: it applies a single shared basis to the compressible
region of each layer and KV head, in contrast to \methodname{}'s per-page
basis. We omit xKV~\citep{chang2025xkv}, which exploits an orthogonal
source of redundancy across layers, while OjaKV~\citep{zhu2026ojakv}
is reported in the supplementary material due to its unstable
reproduced performance across models and tasks.

\ADLinactivate
\begin{table}[!htbp]
\centering
{\scriptsize
\setlength{\tabcolsep}{0.5pt}
\renewcommand{\arraystretch}{0.96}
\begin{tabular}{
  @{}
  >{\centering\arraybackslash}m{0.22in}
  @{\hspace{1pt}}
  >{\raggedright\arraybackslash}m{0.70in}
  @{\hspace{2pt}}|@{\hspace{2pt}}
  *{8}{>{\centering\arraybackslash}m{0.39in}}
  >{\centering\arraybackslash}m{0.55in}
  *{4}{>{\centering\arraybackslash}m{0.39in}}
  @{\hspace{2pt}}|@{\hspace{2pt}}
  >{\centering\arraybackslash}m{0.39in}
  @{}
}
\toprule
\noalign{\vskip 2.5pt}
& \multicolumn{1}{>{\raggedright\arraybackslash}m{0.70in}@{\hspace{2pt}}|@{\hspace{2pt}}}{\multirow{2}{*}{\textbf{Method}}}
& \multicolumn{8}{c}{\textbf{Retrieval}}
& \multicolumn{1}{c}{\textbf{Multi-hop}}
& \multicolumn{2}{c}{\textbf{Agg.}}
& \multicolumn{2}{c@{\hspace{2pt}}|@{\hspace{2pt}}}{\textbf{QA}}
& \multirow{2}{*}{\textbf{Avg.}} \\
&  & \textbf{S1} & \textbf{S2} & \textbf{S3} & \textbf{MK1} &
\textbf{MK2} & \textbf{MK3} & \textbf{MQ} & \textbf{MV} & \textbf{VT} &
\textbf{CWE} & \textbf{FWE} & \textbf{QA-1} &
\textbf{QA-2} & \\
\noalign{\vskip 2.5pt}
\midrule
\multicolumn{16}{c}{RULER 16K} \\
\midrule
\noalign{\vskip 2.5pt}
\multirow{5}{*}{\rotatebox[origin=c]{90}{Llama3.1}}
& Full KV
& 100.00 & 100.00 & 100.00 & 99.60 & 100.00 & 99.00 & 99.90 & 99.00 & 99.80 & 86.52 & 88.87 & 72.07 & 55.60 & 92.33 \\
& Global SVD
& 0.00 & 91.00 & 85.00 & 95.40 & 99.80 & \textbf{98.60} & 46.00 & 34.90 & 0.04 & 50.82 & 93.47 & 66.12 & 52.80 & 62.61 \\
& Palu(G-LRD)
& \textbf{99.80} & \textbf{100.00} & \textbf{99.60} & \textbf{99.60} & 98.60 & 96.00 & 95.60 & 85.65 & 98.32 & 60.04 & 93.13 & 59.02 & 49.00 & 87.26 \\
& H2O
& 92.40 & 51.80 & 33.40 & 83.60 & 76.20 & 36.20 & 55.70 & 59.35 & 97.20 & \textbf{88.66} & \textbf{94.87} & \textbf{71.77} & 52.80 & 68.76 \\
& \textbf{\methodname{}}
& 98.80 & 95.80 & 82.00 & 98.20 & \textbf{100.00} & 93.40 & \textbf{98.85} & \textbf{91.10} & \textbf{98.80} & 84.84 & 88.40 & 70.83 & \textbf{53.60} & \textbf{88.82} \\
\noalign{\vskip 2.5pt}
\midrule
\noalign{\vskip 2.5pt}
\multirow{5}{*}{\rotatebox[origin=c]{90}{Qwen3}}
& Full KV
& 100.00 & 100.00 & 100.00 & 99.60 & 100.00 & 99.60 & 99.85 & 99.65 & 100.00 & 84.24 & 88.07 & 62.98 & 54.40 & 91.41 \\
& Global SVD
& 3.60 & 99.80 & \textbf{99.20} & 99.40 & 99.20 & \textbf{99.00} & 95.20 & 87.45 & 41.40 & 79.52 & 88.33 & 54.52 & 48.00 & 76.51 \\
& Palu(G-LRD)
& \textbf{100.00} & 99.80 & 98.00 & 99.40 & 92.40 & 77.40 & 96.00 & 85.25 & 92.36 & 37.98 & 65.33 & 42.78 & 40.60 & 79.02 \\
& H2O
& 69.40 & 44.80 & 33.40 & 66.60 & 67.20 & 42.00 & 64.65 & 62.50 & 98.72 & 82.54 & \textbf{95.53} & 55.62 & 49.00 & 64.00 \\
& \textbf{\methodname{}}
& \textbf{100.00} & \textbf{100.00} & 99.00 & \textbf{99.60} & \textbf{99.80} & 98.40 & \textbf{99.90} & \textbf{99.60} & \textbf{99.96} & \textbf{83.78} & 86.67 & \textbf{62.22} & \textbf{50.60} & \textbf{90.73} \\
\noalign{\vskip 2.5pt}
\midrule
\multicolumn{16}{c}{RULER 32K} \\
\midrule
\noalign{\vskip 2.5pt}
\multirow{5}{*}{\rotatebox[origin=c]{90}{Llama3.1}}
& Full KV
& 100.00 & 100.00 & 100.00 & 99.80 & 99.40 & 99.60 & 99.90 & 98.35 & 98.92 & 26.38 & 93.53 & 70.88 & 51.20 & 87.54 \\
& Global SVD
& 0.00 & 84.20 & 77.20 & 95.80 & 99.00 & \textbf{99.00} & 59.60 & 46.95 & 0.00 & 1.60 & \textbf{95.00} & 64.37 & 49.20 & 59.38 \\
& Palu(G-LRD)
& \textbf{100.00} & \textbf{98.40} & \textbf{98.60} & 99.00 & 96.40 & 94.20 & 95.45 & 91.20 & 94.96 & 24.12 & 81.07 & 57.02 & 44.40 & 82.68 \\
& H2O
& 99.80 & 47.80 & 31.60 & 80.80 & 68.20 & 40.00 & 56.25 & 51.80 & 96.32 & \textbf{24.32} & 90.00 & 66.92 & 49.00 & 61.75 \\
& \textbf{\methodname{}}
& 99.80 & 97.60 & 83.00 & \textbf{99.60} & \textbf{99.80} & 95.00 & \textbf{98.85} & \textbf{91.40} & \textbf{98.28} & 23.04 & 93.13 & \textbf{67.38} & \textbf{50.20} & \textbf{84.39} \\
\noalign{\vskip 2.5pt}
\midrule
\noalign{\vskip 2.5pt}
\multirow{5}{*}{\rotatebox[origin=c]{90}{Qwen3}}
& Full KV
& 100.00 & 100.00 & 100.00 & 99.60 & 97.20 & 94.80 & 99.65 & 98.75 & 99.96 & 64.36 & 96.80 & 59.82 & 51.80 & 89.44 \\
& Global SVD
& 0.00 & \textbf{100.00} & \textbf{99.80} & 99.00 & 96.60 & \textbf{94.00} & 98.85 & \textbf{98.50} & 7.16 & 48.06 & 94.67 & 50.53 & 47.40 & 71.89 \\
& Palu(G-LRD)
& 99.80 & \textbf{100.00} & 98.40 & 96.20 & 75.00 & 45.20 & 93.40 & 80.95 & 83.20 & 18.44 & 55.20 & 37.95 & 39.60 & 71.03 \\
& H2O
& 97.40 & 39.40 & 31.60 & 62.00 & 64.60 & 42.20 & 61.20 & 52.60 & 99.16 & 61.98 & \textbf{96.73} & \textbf{59.42} & \textbf{50.00} & 62.95 \\
& \textbf{\methodname{}}
& \textbf{100.00} & 99.80 & 98.60 & \textbf{99.80} & \textbf{97.00} & 84.60 & \textbf{99.60} & 98.35 & \textbf{99.76} & \textbf{62.80} & 95.93 & 54.18 & 47.40 & \textbf{87.53} \\
\noalign{\vskip 2.5pt}
\bottomrule
\end{tabular}
}
\caption{RULER performance of Llama-3.1-8B-Instruct and Qwen3-8B at
16K and 32K context lengths under a KV cache storage budget of 60\%.}
\label{tab:ruler-results}
\end{table}
\ADLactivate

\paragraph{Implementation.}
We implement \methodname{} and Global SVD in Hugging Face
Transformers~\citep{wolf2020transformers} and evaluate all methods with
lm-eval-harness~\citep{gao2024lm_eval_harness} on NVIDIA GH200 GPUs.
Llama-3.1-8B-Instruct uses its default chat template, and all evaluations
use greedy decoding, so the reported accuracies are deterministic and
obtained from a single run. Unless otherwise stated, \methodname{} uses a
page size of $P=32$ and ranks $(r_K,r_V)=(16,14)$, giving a KV cache
storage ratio of approximately 0.6. The uncompressed dense region
consists of one sink page (32 tokens) together with the most recently
completed page and the current incomplete page.

\ADLinactivate
\begin{table}[!htbp]
\centering
{\footnotesize
\setlength{\tabcolsep}{3pt}
\renewcommand{\arraystretch}{1.00}
\begin{tabular}{
  @{}
  >{\centering\arraybackslash}m{0.16in}
  >{\raggedright\arraybackslash}m{0.78in}
  @{\hspace{0.5pt}}|@{\hspace{0.5pt}}
  >{\centering\arraybackslash}m{0.84in}
  @{\hspace{1.5pt}}
  >{\centering\arraybackslash}m{0.84in}
  >{\centering\arraybackslash}m{0.42in}
  >{\centering\arraybackslash}m{0.60in}
  >{\centering\arraybackslash}m{0.62in}
  >{\centering\arraybackslash}m{0.42in}
  @{\hspace{0.5pt}}|@{\hspace{0.5pt}}
  >{\centering\arraybackslash}m{0.42in}
  @{}
}
\toprule
\noalign{\vskip 2.5pt}
\multicolumn{2}{@{}l@{\hspace{0.5pt}}|@{\hspace{0.5pt}}}{\textbf{Method}}
& \textbf{Single-Doc QA}
& \textbf{Multi-Doc QA}
& \textbf{Sum.}
& \textbf{Few-shot}
& \textbf{Synthetic}
& \textbf{Code}
& \textbf{Avg.} \\
\noalign{\vskip 2.5pt}
\midrule
\noalign{\vskip 2.5pt}
\multirow{5}{*}{\rotatebox[origin=c]{90}{Llama3.1}}
& Full KV & 43.47 & 46.14 & 29.00 & 69.32 & 54.25 & 30.45 & 45.83 \\
& Global SVD & 42.79 & 45.13 & 26.83 & 66.86 & 53.80 & 27.86 & 44.26 \\
& Palu & 40.39 & 42.83 & 26.98 & 68.39 & 46.50 & 23.31 & 42.21 \\
& H2O & 42.99 & \textbf{45.78} & \textbf{28.41} & \textbf{69.02} &
\textbf{55.25} & \textbf{30.34} & \textbf{45.61} \\
& \textbf{\methodname{}} & \textbf{43.34} & 45.58 & 27.82 & 68.50 & 54.50 & 30.32 & 45.33 \\
\noalign{\vskip 2.5pt}
\midrule
\noalign{\vskip 2.5pt}
\multirow{5}{*}{\rotatebox[origin=c]{90}{Qwen3}}
& Full KV & 13.97 & 11.12 & 24.62 & 69.01 & 49.57 & 11.23 & 29.86 \\
& Global SVD & 12.98 & 10.19 & 23.17 & \textbf{68.91} & 44.38 & 10.63 & 28.49 \\
& Palu & 9.63 & 7.87 & 23.69 & 64.99 & 35.95 & 4.92 & 25.02 \\
& H2O & 13.20 & 10.78 & \textbf{24.45} & 68.67 & 46.09 & 11.07 & 29.10 \\
& \textbf{\methodname{}} & \textbf{13.51} & \textbf{10.98} & 23.27 & 68.39 &
\textbf{50.41} & \textbf{11.18} & \textbf{29.48} \\
\noalign{\vskip 2.5pt}
\bottomrule
\end{tabular}
}
\caption{Category-level LongBench results (\%) at $\rho=0.6$.
Complete per-task results are provided in the supplementary material.}
\label{tab:longbench-results}
\end{table}
\ADLactivate

\subsection{Main Results}
We evaluate \methodname{} and the baselines on RULER and LongBench under a
matched budget of $\rho\approx 0.6$. \methodname{} compresses each completed
page and keeps only a few dense sink and recent pages; ignoring these,
its factorized-page storage ratio is
$\rho_{\text{\methodname{}}}=\frac{(r_K+r_V)(P+d)}{2Pd}\approx 0.586$ at
$(r_K,r_V)=(16,14)$, $P=32$, and head dimension $d=128$ (shared by both
models). Global SVD instead shares a single basis over the compressible
region of each layer and KV head: for an input of length $S$ it stores
$\rho_{\mathrm{global}}=\frac{r(S+d)}{Sd}$, and we select $r$ per input
to match the budget, giving ranks of 76--78. 
Both \methodname{} and Global SVD decompose post-RoPE keys,
which avoids extra RoPE transformations~\citep{su2024roformer} at cache
access; pre-RoPE keys are slightly more compressible but give only a
small accuracy gain in our ablation. Complete results and
configurations are provided in the supplementary material.
For Palu, we use the
official G-LRD variant (four-head groups, Fisher-uniform rank
allocation, no quantization); for H2O, we follow its equal-split policy,
allocating $0.3$ of the KV budget to heavy-hitter tokens and $0.3$ to
recent tokens.

\paragraph{Results on RULER.}
Table~\ref{tab:ruler-results} reports RULER results at input lengths of
16K and 32K. \methodname{} achieves the highest average among all compressed
methods in every model--context setting.
On Llama3.1, \methodname{} obtains 88.82 and 84.39 at 16K and 32K,
corresponding to 96.2\% and 96.4\% of Full KV performance. It leads
Global SVD by 26.21 and 25.01 points and H2O by 20.06 and 22.64 points,
while remaining competitive with Palu (within 1.56 and 1.71 points).
On Qwen3, \methodname{} obtains 90.73 and 87.53 at 16K and 32K, corresponding
to 99.3\% and 97.9\% of Full KV performance. It leads Global SVD by
14.22 and 15.64 points, H2O by 26.73 and 24.58 points, and Palu by 11.71
and 16.50 points. 
Overall, page-wise low-rank compression
delivers strong and consistent accuracy across the evaluated models,
context lengths, and task types.

\paragraph{Results on LongBench.}
Table~\ref{tab:longbench-results} reports category-level LongBench
results. \methodname{} scores 45.33 on Llama3.1 and 29.48 on Qwen3, corresponding to
98.9\% and 98.7\% of Full KV performance. Among compressed methods it
ranks first on Qwen3, leading H2O by 0.38 points, and second on Llama3.1,
within 0.28 points of H2O; it further outperforms Global SVD by 1.07 and
0.99 points and Palu by 3.12 and 4.46 points on Llama3.1 and Qwen3,
respectively. These results show strong cross-model accuracy on
real-world tasks and a comparable or consistent advantage over the other low-rank baselines.

\paragraph{Observations.}
Because RULER consists of controlled synthetic tasks that isolate
specific capabilities more cleanly than the real-world tasks in
LongBench, we examine the tasks on which \methodname{} achieves especially
large gains to characterize the benefits of page-wise compression.

\textbf{\textit{Page-wise versus sequence-level bases.}}
\methodname{} shows a pronounced advantage on NIAH-S1 and Variable Tracking
(VT), where a sequence-level basis collapses. NIAH-S1 hides a unique
key--value pair in a highly repetitive context~\citep{hsieh2024ruler}:
at 16K / 32K, \methodname{} scores 98.80 / 99.80 on Llama3.1 and 100.00 / 100.00 on
Qwen3, whereas Global SVD scores 0.00 / 0.00 and 3.60 / 0.00. VT requires
recovering a five-variable, four-hop assignment chain among repeated
distractors: \methodname{} scores 98.80 / 98.28 on Llama3.1 and 99.96 / 99.76 on
Qwen3, versus 0.04 / 0.00 and 41.40 / 7.16 for Global SVD. In both cases,
\textit{page-local bases preserve the sparse or intermediate content that a
sequence-level objective discards once repeated background dominates its
reconstruction}.

\textbf{\textit{Compression versus eviction.}}
\methodname{} also holds a clear advantage on NIAH-S3, which requires exact
recovery of a long UUID. At 16K / 32K it scores 82.00 / 83.00 on Llama3.1
and 99.00 / 98.60 on Qwen3, whereas H2O scores 33.40 / 31.60 on both models.
\textit{Because \methodname{} retains every token in compressed form, the full UUID
stays available to later queries}; H2O, having irreversibly evicted the
relevant KV entries, frequently emits incomplete or corrupted UUIDs.

\FloatBarrier
\subsection{Quantization Compatibility}
\label{sec:quantization-results}

We further demonstrate \methodname{}'s compatibility with low-bit
quantization, following the broader success of post-training low-bit
quantization~\citep{xiao2023smoothquant}. For each truncated decomposition,
we absorb the singular values
$\Sigma$ into the two page-wise factors, constructing
$
    \widetilde{L}_{X,p}
    =
    U_{X,p}\Sigma_{X,p}^{1/2},
    \widetilde{R}_{X,p}
    =
    \Sigma_{X,p}^{1/2}V_{X,p}^{\top}.
$
and apply symmetric INT4
quantization to each factor matrix: 
column-wise for
$\widetilde{L}_{K,p}$ and $\widetilde{R}_{K,p}$, row-wise for
$\widetilde{L}_{V,p}$, and column-wise for $\widetilde{R}_{V,p}$.
For the direct INT3 and INT4 baselines, we apply per-channel quantization to
keys and per-token quantization to values within each uncompressed KV page~\citep{liu2024kivi}
while keeping the same dense sink and local regions.

Table~\ref{tab:quantization-compatibility} shows that combining \methodname{}
with INT4 reduces the KV storage ratio to 0.187 while staying within
2.58 points on Llama3.1 and 0.53 points on Qwen3 of the unquantized
\methodname{} results. At this smallest budget, it remains competitive with
direct INT3 on Full KV (storage ratio 0.223) despite a smaller budget,
staying close on Qwen3 (90.20 vs.\ 91.02) while trailing on Llama3.1
(86.24 vs.\ 88.51). Direct INT4 (storage ratio 0.283) attains higher
accuracy but at a larger budget. These results show that factor
quantization composes with page-wise low-rank compression to reach an
aggressive 0.187 storage regime with limited accuracy loss.

\ADLinactivate
\begin{table}[!htbp]
\centering
{\footnotesize
\setlength{\tabcolsep}{5pt}
\renewcommand{\arraystretch}{1.05}
\begin{tabular}{@{}l@{\hspace{3pt}}|@{\hspace{3pt}}ccc@{}}
\toprule
\noalign{\vskip 2.5pt}
\textbf{Method} &
\textbf{Storage} & \textbf{Llama3.1} & \textbf{Qwen3} \\
\noalign{\vskip 2.5pt}
\midrule
\noalign{\vskip 2.5pt}
Full KV          & 1.000 & 92.33 & 91.41 \\
Full KV + INT4   & 0.283 & 92.34 & 91.09 \\
Full KV + INT3   & 0.223 & 88.51 & 91.02 \\
\noalign{\vskip 2.5pt}
\noalign{\hbox to 2.60in{%
  \leaders\hbox{\rule{2pt}{0.35pt}\hskip 2pt}\hfill\kern0pt}}
\noalign{\vskip 2.5pt}
\methodname{}         & 0.588 & 88.82 & 90.73 \\
\methodname{} + INT4  & 0.187 & 86.24 & 90.20 \\
\noalign{\vskip 2.5pt}
\bottomrule
\end{tabular}
}
\caption{Average RULER scores (\%) at 16K input length under different precision settings.}
\label{tab:quantization-compatibility}
\end{table}
\ADLactivate

\subsection{Ablation Study}

We conduct all ablation studies on RULER using
Llama3.1 with a 16K input length, as this setting more
clearly differentiates among compression methods.

\begin{figure}[!htbp]
\centering
\includegraphics[width=0.56\textwidth]{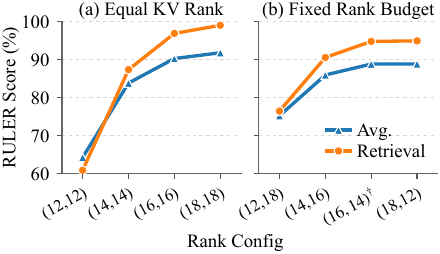}
\caption{Overall and Retrieval scores over different KV rank configurations.
$\dagger$ denotes the configuration used in the main experiment.}
\label{fig:rank-selection}
\end{figure}

\paragraph{Rank Selection.}
We examine how the total rank budget and its allocation between keys
and values affect downstream performance. As shown in
Figure~\ref{fig:rank-selection}(a), increasing the shared K/V rank
consistently improves performance, with diminishing gains beyond rank
16. Figure~\ref{fig:rank-selection}(b) compares different allocations
under the fixed budget $r_K+r_V=30$. The $(16,14)$ configuration
outperforms $(14,16)$ in both the overall and Retrieval averages,
indicating that an insufficient key rank cannot be offset by assigning
more rank to values. Increasing $r_K$ from 16 to 18 provides little
additional improvement in the overall average. We therefore use
$(r_K,r_V)=(16,14)$ as the default configuration, allocating slightly
more rank to keys without increasing the total storage budget. Complete
ablation results are provided in the supplementary material.

\paragraph{Page Size Selection.}

\begin{figure}[!htbp]
\centering
\includegraphics[width=0.38\textwidth]{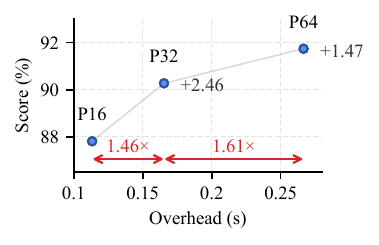}
\caption{Full RULER average score and decomposition overhead during
prefilling across page sizes (KV cache budget $\rho=0.6$).}
\label{fig:page-size-selection}
\end{figure}

Figure~\ref{fig:page-size-selection} compares different page sizes
under the same compression ratio.
At the same compression ratio, \methodname{} remains effective across all
evaluated page sizes. As the page size grows from 16 to 32, accuracy
improves substantially. However, when expand the page size to 64, it brings a smaller gain but
noticeably higher decomposition latency. 
 We therefore use $P=32$ in all experiments.

\subsection{System Efficiency}

We evaluate \methodname{} on Qwen3 using sampled NIAH-S1
inputs~\citep{hsieh2024ruler}. Raw KV uses the same paged KV cache
implementation as \methodname{} but without compression, allowing us to
measure the overhead introduced specifically by \methodname{}. We also
include Hugging Face StaticCache~\citep{wolf2020transformers} to
quantify the overhead of our paged KV cache implementation. All latency
measurements use a batch size of one, with the PyTorch SDPA
FlashAttention backend explicitly enabled for dense attention.

\begin{figure}[!htbp]
\centering
\includegraphics[width=0.48\textwidth]{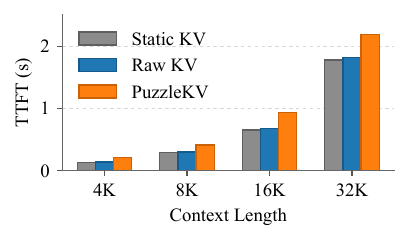}
\caption{TTFT of Static KV, Raw KV, and \methodname{} across different
context lengths.}
\label{fig:runtime-latency}
\end{figure}

\paragraph{Runtime Latency.}
All methods use the same warm-up schedule. Figure~\ref{fig:runtime-latency}
compares the time to first token(TTFT) of \methodname{} with Static KV and Raw KV.
At 32K,
\methodname{} introduces a 20.9\% TTFT overhead relative to Raw KV. As
shown in Table~\ref{tab:runtime-tpot}, its time per output token(TPOT) is only
0.075 ms/token (0.18\%) higher than Raw KV. The runtime cost of
\methodname{} is therefore concentrated in page decomposition during
prefill, with negligible additional overhead during decoding.

\begin{table}[!htbp]
\centering
{\small
\setlength{\tabcolsep}{5pt}
\renewcommand{\arraystretch}{1.05}
\begin{tabular}{@{}lc@{}}
\toprule
\textbf{Method} & \textbf{TPOT (ms/token)} \\
\midrule
Static KV & 40.477 \\
Raw KV & 42.662 \\
\methodname{} & 42.737 \\
\bottomrule
\end{tabular}
}
\caption{TPOT at 16K context length with 256 generated
tokens.}
\label{tab:runtime-tpot}
\end{table}

\paragraph{Memory Footprint.}
Table~\ref{tab:logical-live-kv-footprint} reports the logical live KV
cache footprint at a 16K context length with 256 generated tokens.
At steady state, \methodname{} uses 58.76\% of the Raw KV footprint, only
0.17 percentage points above the theoretical page-level ratio.
Incremental page conversion adds just 12.45 MiB at peak.

\begin{table}[!htbp]
\centering
{\small
\setlength{\tabcolsep}{3pt}
\renewcommand{\arraystretch}{1.05}
\begin{tabular}{@{}l@{\hspace{6.4pt}}cc@{}}
\toprule
\noalign{\vskip 2.5pt}
\textbf{Stage}
  & \textbf{Raw KV (MiB)} & \textbf{\methodname{} (MiB)} \\
\noalign{\vskip 2.5pt}
\midrule
\noalign{\vskip 2.5pt}
Prefill              & 2204.85 & 2204.85 \\
Initial compression  & 2204.85 & 1297.05 \\
Update peak          & 2205.90 & 1309.50 \\
Update complete      & 2205.90 & 1296.25 \\
\noalign{\vskip 2.5pt}
\bottomrule
\end{tabular}
}
\caption{Logical live KV cache footprint across execution stages.}
\label{tab:logical-live-kv-footprint}
\end{table}

\FloatBarrier
\section{Discussion}
Beyond accuracy, organizing compression around the page 
aligns it with the fixed-size cache blocks of modern serving engines
and enables regular, batched GPU
decomposition. Our current implementation is a batch-one standalone
prototype; because it shares this page-based abstraction, it has strong
potential for integration into a batched paged serving engine such as
vLLM~\citep{kwon2023efficient}.

\section{Conclusion}

We presented \methodname{}, a training- and calibration-free method for
page-wise low-rank KV cache compression. \methodname{} combines batched
page decomposition, direct attention over dense and factorized pages,
and incremental conversion during decoding.
Experiments across models, context lengths, and benchmarks show that
page-wise decomposition achieves strong accuracy under constrained KV
cache budgets, substantially outperforms sequence-level low-rank
compression, and can be combined with low-bit quantization to reach more
aggressive storage budgets.
More
broadly, \methodname{} shows that compression granularity can jointly shape
representation quality, cache management, and GPU execution.

\bibliographystyle{plainnat}
\bibliography{references}

\clearpage
\appendix
\begin{center}
  {\LARGE\bfseries Appendix\par}
\end{center}
\vspace{0.75em}
\input{appendix}

\end{document}

%% file: appendix.tex
\section{Experimental Setup}
\label{app:experimental-details}

\subsection{Additional Evaluation Details}

All benchmark evaluations use a fixed random seed of 42. RULER uses the
lm-eval 0.4.12 implementation and contains 13 tasks with 500 examples per
task. LongBench uses the \texttt{test} split of \texttt{THUDM/LongBench};
Appendix~\ref{app:longbench-results} additionally reports the five Chinese
tasks omitted from the main table. Qwen3 is evaluated without a chat
template and with thinking disabled. For LongBench, Llama-3.1-8B-Instruct
uses raw prompts for six tasks and its default chat template for the other
15 tasks.

\subsection{Computing Environment}

\vspace{4pt}
\begin{center}
\begin{minipage}{0.50\textwidth}
\centering
\footnotesize
\renewcommand{\arraystretch}{1.12}
\setlength{\tabcolsep}{3.2pt}
\begin{tabular}{@{}
>{\raggedright\arraybackslash}m{0.28\columnwidth}
>{\raggedright\arraybackslash}m{0.58\columnwidth}
@{}}
\toprule
\textbf{Setting} & \textbf{Accuracy experiments} \\
\midrule
GPU
& 1$\times$ NVIDIA GH200 (120\,GB) \\
CPU
& 72-core ARM Neoverse-V2 \\
System memory
& 212.70\,GiB \\
OS
& Rocky Linux 9.7 \\
Python
& 3.11.8 \\
CUDA
& 12.8 \\
PyTorch
& 2.9.1+\texttt{cu128} \\
Transformers
& 5.10.2 \\
Triton
& 3.5.1 \\
\bottomrule
\end{tabular}
\captionof{table}{Computing environment for accuracy experiments. The
system-efficiency study uses the same configuration except for CUDA 13.0 and
PyTorch 2.12.0+\texttt{cu130}, because the accuracy environment exhibits an
SVD performance issue.}
\label{tab:computing-environments}
\end{minipage}
\end{center}

\subsection{Baseline Configurations}

\paragraph{Global SVD.}
Global SVD uses the same dense sink and local-page policy as PuzzleKV.
Its input-dependent rank is selected by accounting for both its low-rank
factors and these dense regions when matching the target storage budget.

\paragraph{Palu.}
Palu uses the \texttt{Salesforce/wikitext}
\texttt{wikitext-2-raw-v1} training split for calibration. Fisher estimation uses
32 sequences of 1,024 tokens, and whitening uses 256 sequences of
1,024 tokens, both with seed 42. Due to rank alignment, the realized
retained parameter ratio rounds to 0.60 for both models.

\paragraph{H2O.}
The H2O budget is computed from each tokenized prompt length. Selection is
performed independently per layer and KV head using accumulated post-softmax
attention mass. For GQA models, attention mass from query heads mapped to
the same KV head is summed before selection. Prefill uses full attention,
after which H2O evicts entries from the completed prompt cache before
autoregressive decoding.

\paragraph{OjaKV.}
We select OjaKV-PF, the best-performing reported variant, because its
full-attention prefill followed by compressed decoding matches PuzzleKV's
high-level execution pattern; standard OjaKV instead applies compression during
prefill. We evaluate OjaKV-PF at the same nominal retained KV cache ratio of
0.60. We use rank
$r=77$ for both keys and values, initialize the basis by SVD on
the same WikiText-2 training split with calibration seed 2, and update the
basis online during evaluation. Our implementation samples at most 256
pooled rows for each online update.

\section{Additional Evaluation Results}

\subsection{Complete Quantization Compatibility Results}
\label{app:quantization-results}

Table~\ref{tab:quantization-results-full} reports all 13 RULER tasks at
16K; PuzzleKV uses $P=32$ and $(r_K,r_V)=(16,14)$.

\begin{center}
\begin{minipage}{\textwidth}
\centering
\footnotesize
\renewcommand{\arraystretch}{1.14}
\setlength{\tabcolsep}{0.85pt}
\begin{tabular*}{\textwidth}{@{\extracolsep{\fill}}lc*{14}{c}@{}}
\toprule
\multicolumn{2}{c}{}
& \multicolumn{8}{c}{\textbf{Retrieval}}
& \multicolumn{1}{c}{\textbf{Multi-hop}}
& \multicolumn{2}{c}{\textbf{Agg.}}
& \multicolumn{2}{c}{\textbf{QA}}
& \multicolumn{1}{c}{} \\
\cmidrule(lr){3-10}
\cmidrule(lr){11-11}
\cmidrule(lr){12-13}
\cmidrule(lr){14-15}
\textbf{Method} & \textbf{Storage}
& \textbf{S1} & \textbf{S2} & \textbf{S3}
& \textbf{MK1} & \textbf{MK2} & \textbf{MK3}
& \textbf{MQ} & \textbf{MV} & \textbf{VT}
& \textbf{CWE} & \textbf{FWE}
& \textbf{QA-1} & \textbf{QA-2} & \textbf{Avg.} \\
\midrule
\multicolumn{16}{@{}l}{\textit{Llama-3.1-8B-Instruct}} \\
Full KV & 100.00 & 100.00 & 100.00 & 100.00 & 99.60 & 100.00 & 99.00 & 99.90 & 99.00 & 99.80 & 86.52 & 88.87 & 72.07 & 55.60 & 92.33 \\
Full KV + INT4 & 28.30 & 100.00 & 100.00 & 100.00 & 99.60 & 100.00 & 99.20 & 99.90 & 98.30 & 99.56 & 87.28 & 88.40 & 72.57 & 55.60 & 92.34 \\
Full KV + INT3 & 22.30 & 100.00 & 100.00 & 99.60 & 99.20 & 98.20 & 87.40 & 98.20 & 91.45 & 95.04 & 70.48 & 87.00 & 69.90 & 54.20 & 88.51 \\
\noalign{\vskip 1.2pt}\hdashline\noalign{\vskip 1.2pt}
\textbf{PuzzleKV} & 58.80 & 98.80 & 95.80 & 82.00 & 98.20 & 100.00 & 93.40 & 98.85 & 91.10 & 98.80 & 84.84 & 88.40 & 70.83 & 53.60 & 88.82 \\
\textbf{PuzzleKV + INT4} & 18.70 & 98.00 & 93.80 & 74.00 & 97.40 & 99.80 & 89.20 & 96.80 & 80.25 & 94.40 & 84.48 & 88.00 & 70.93 & 54.00 & 86.24 \\
\midrule
\multicolumn{16}{@{}l}{\textit{Qwen3-8B}} \\
Full KV & 100.00 & 100.00 & 100.00 & 100.00 & 99.60 & 100.00 & 99.60 & 99.85 & 99.65 & 100.00 & 84.24 & 88.07 & 62.98 & 54.40 & 91.41 \\
Full KV + INT4 & 28.30 & 100.00 & 100.00 & 100.00 & 99.60 & 100.00 & 99.40 & 99.90 & 99.50 & 100.00 & 84.54 & 86.93 & 62.55 & 51.80 & 91.09 \\
Full KV + INT3 & 22.30 & 100.00 & 100.00 & 100.00 & 99.60 & 99.40 & 98.80 & 99.60 & 99.40 & 99.88 & 81.02 & 91.53 & 62.08 & 52.00 & 91.02 \\
\noalign{\vskip 1.2pt}\hdashline\noalign{\vskip 1.2pt}
\textbf{PuzzleKV} & 58.80 & 100.00 & 100.00 & 99.00 & 99.60 & 99.80 & 98.40 & 99.90 & 99.60 & 99.96 & 83.78 & 86.67 & 62.22 & 50.60 & 90.73 \\
\textbf{PuzzleKV + INT4} & 18.70 & 100.00 & 99.80 & 98.60 & 99.80 & 99.80 & 92.40 & 99.65 & 99.65 & 99.88 & 82.76 & 88.40 & 59.82 & 52.00 & 90.20 \\
\bottomrule
\end{tabular*}
\captionof{table}{Complete quantization-compatibility results (\%) on
RULER at 16K. Storage is reported as a percentage of Full KV storage.}
\label{tab:quantization-results-full}
\end{minipage}
\end{center}

\subsection{Complete LongBench Results}
\label{app:longbench-results}

Table~\ref{tab:full-longbench-results} gives per-task scores for the
14 English tasks and two code tasks used in the main paper.
Table~\ref{tab:longbench-chinese-results} adds the five Chinese tasks.

\begin{center}
\begin{minipage}{\textwidth}
\centering
\footnotesize
\setlength{\tabcolsep}{1.20pt}
\begin{tabular*}{\textwidth}{@{\extracolsep{\fill}}l*{17}{c}@{}}
\toprule
\multicolumn{1}{c}{}
& \multicolumn{3}{c}{\textbf{Single-Doc QA}}
& \multicolumn{3}{c}{\textbf{Multi-Doc QA}}
& \multicolumn{3}{c}{\textbf{Summarization}}
& \multicolumn{3}{c}{\textbf{Few-shot}}
& \multicolumn{2}{c}{\textbf{Synthetic}}
& \multicolumn{2}{c}{\textbf{Code}}
& \multicolumn{1}{c}{} \\
\cmidrule(lr){2-4}
\cmidrule(lr){5-7}
\cmidrule(lr){8-10}
\cmidrule(lr){11-13}
\cmidrule(lr){14-15}
\cmidrule(lr){16-17}
\textbf{Method} & \textbf{NQA} & \textbf{Qasp} &
\textbf{MF-en} & \textbf{HQA} & \textbf{2WQA} & \textbf{Mus} &
\textbf{GRep} & \textbf{QMS} & \textbf{MNews} & \textbf{TREC} &
\textbf{TQA} & \textbf{SAM} & \textbf{PC} & \textbf{PR-en} &
\textbf{LCC} & \textbf{RB-P} & \textbf{Avg.} \\
\midrule
\multicolumn{18}{@{}l}{\textit{Llama-3.1-8B-Instruct}} \\
Full KV & 29.51 & 44.98 & 55.92 & 57.53 & 48.09 & 32.81 & 34.61 & 25.50 & 26.89 & 72.50 & 92.12 & 43.33 & 9.00 & 99.50 & 29.42 & 31.47 & 45.83 \\
Global SVD & 29.97 & \textbf{44.18} & 54.21 & 55.61 & 46.39 & \textbf{33.40} & 30.37 & 24.11 & 26.00 & 69.50 & 91.71 & 39.36 & 9.60 & 98.00 & 28.42 & 27.30 & 44.26 \\
Palu (G-LRD) & 26.84 & 42.22 & 52.10 & 50.89 & 46.49 & 31.11 & 30.73 & 24.55 & 25.67 & \textbf{73.00} & 88.46 & \textbf{43.70} & 0.50 & 92.50 & 18.45 & 28.17 & 42.21 \\
H2O & 29.88 & 43.23 & 55.86 & 57.48 & 47.06 & 32.81 & \textbf{33.74} & \textbf{25.48} & 26.02 & 72.00 & 91.67 & 43.38 & \textbf{11.00} & 99.50 & 29.06 & \textbf{31.62} & \textbf{45.61} \\
\textbf{PuzzleKV} & \textbf{30.10} & 43.19 & \textbf{56.72} & \textbf{57.77} & \textbf{47.40} & 31.56 & 31.97 & 25.42 & \textbf{26.08} & 71.00 & \textbf{92.05} & 42.45 & 9.00 & \textbf{100.00} & \textbf{29.46} & 31.18 & 45.33 \\
\midrule
\multicolumn{18}{@{}l}{\textit{Qwen3-8B}} \\
Full KV & 3.71 & 11.48 & 26.72 & 12.77 & 12.96 & 7.63 & 29.76 & 21.15 & 22.94 & 71.50 & 90.71 & 44.83 & 7.14 & 92.00 & 11.20 & 11.26 & 29.86 \\
Global SVD & 3.51 & 11.44 & 23.98 & 11.53 & 11.81 & 7.24 & 28.64 & 18.12 & 22.76 & \textbf{72.00} & 90.71 & 44.02 & 2.25 & 86.50 & 10.71 & 10.55 & 28.49 \\
Palu (G-LRD) & 2.87 & 10.54 & 15.49 & 9.78 & 7.69 & 6.14 & 27.71 & \textbf{22.85} & 20.52 & 67.00 & 84.24 & 43.72 & 3.72 & 68.17 & 4.61 & 5.23 & 25.02 \\
H2O & 3.77 & 10.45 & \textbf{25.37} & 12.56 & \textbf{12.72} & 7.07 & \textbf{29.69} & 20.78 & \textbf{22.87} & 70.50 & 90.96 & \textbf{44.54} & 6.34 & 85.83 & \textbf{11.31} & 10.82 & 29.10 \\
\textbf{PuzzleKV} & \textbf{3.95} & \textbf{11.61} & 24.97 & \textbf{12.60} & 12.65 & \textbf{7.70} & 28.33 & 18.99 & 22.50 & 71.00 & \textbf{91.21} & 42.96 & \textbf{6.37} & \textbf{94.45} & 11.15 & \textbf{11.20} & \textbf{29.48} \\
\bottomrule
\end{tabular*}
\captionof{table}{Complete per-task LongBench results on the English-language and code
tasks summarized in the main paper. All compressed methods use the same KV
cache storage ratio $\rho=0.60$.}
\label{tab:full-longbench-results}
\end{minipage}
\end{center}

\begin{center}
\begin{minipage}{0.90\textwidth}
\centering
\footnotesize
\begin{tabular*}{\linewidth}{@{\extracolsep{\fill}}lccccccc@{}}
\toprule
\textbf{Method} & \textbf{MF-zh} & \textbf{DuR} & \textbf{VCSum}
& \textbf{LSHT} & \textbf{PR-zh} & \textbf{Avg.-16} & \textbf{Avg.-21} \\
\midrule
\multicolumn{8}{@{}l}{\textit{Llama-3.1-8B-Instruct}} \\
Full KV & 62.82 & 34.36 & 17.07 & 46.00 & 90.45 & 45.83 & 46.85 \\
Global SVD & 61.26 & 27.96 & 16.60 & 44.00 & 83.68 & 44.26 & 44.84 \\
Palu (G-LRD) & 55.70 & 28.55 & 15.15 & \textbf{46.50} & 71.00 & 42.21 & 42.49 \\
H2O & 61.57 & \textbf{32.79} & \textbf{17.42} & 43.00 & 91.39 & \textbf{45.61} & \textbf{46.47} \\
\textbf{PuzzleKV} & \textbf{61.66} & 30.25 & 16.31 & 42.00 & \textbf{98.00} & 45.33 & 46.36 \\
\midrule
\multicolumn{8}{@{}l}{\textit{Qwen3-8B}} \\
Full KV & 20.55 & 22.79 & 14.35 & 47.50 & 97.60 & 29.86 & 32.41 \\
Global SVD & 20.58 & 21.56 & \textbf{15.17} & \textbf{46.75} & \textbf{97.67} & 28.49 & 31.31 \\
Palu (G-LRD) & 15.33 & 21.01 & 11.38 & 38.50 & 80.46 & 25.02 & 27.00 \\
H2O & 19.92 & \textbf{23.95} & 14.44 & 44.50 & 91.10 & 29.10 & 31.40 \\
\textbf{PuzzleKV} & \textbf{20.65} & 21.63 & 13.32 & 45.00 & 97.63 & \textbf{29.48} & \textbf{31.90} \\
\bottomrule
\end{tabular*}
\captionof{table}{Additional LongBench results for the five Chinese tasks.
Avg.-16 covers the 14 English-language and two code tasks; Avg.-21 covers
the complete benchmark. All compressed methods use $\rho=0.60$.}
\label{tab:longbench-chinese-results}
\end{minipage}
\end{center}

\section{OjaKV-PF Reproduction Results}
\label{app:ojakv-results}

We evaluate the OjaKV-PF variant at the same nominal retained KV cache ratio
$\rho=0.60$ used by the compressed methods in the main evaluation.

\begin{center}
\begin{minipage}{\textwidth}
Table~\ref{tab:ojakv-ruler-16k} reports the 16K RULER results; zero entries
are measured scores rather than missing results.
\medskip

\centering
\footnotesize
\renewcommand{\arraystretch}{1.12}
\setlength{\tabcolsep}{3.4pt}
\begin{tabular}{@{}l*{13}{c}c@{}}
\toprule
\textbf{Model}
& \multicolumn{8}{c}{\textbf{Retrieval}}
& \multicolumn{1}{c}{\textbf{Multi-hop}}
& \multicolumn{2}{c}{\textbf{Agg.}}
& \multicolumn{2}{c}{\textbf{QA}}
& \textbf{Avg.} \\
\cmidrule(lr){2-9}
\cmidrule(lr){10-10}
\cmidrule(lr){11-12}
\cmidrule(lr){13-14}
& \textbf{S1} & \textbf{S2} & \textbf{S3}
& \textbf{MK1} & \textbf{MK2} & \textbf{MK3}
& \textbf{MQ} & \textbf{MV}
& \textbf{VT} & \textbf{CWE} & \textbf{FWE}
& \textbf{QA-1} & \textbf{QA-2} & \\
\midrule
Llama-3.1-8B-Instruct
& 50.80 & 99.40 & 97.20
& 99.20 & 98.80 & 51.80
& 37.25 & 68.25
& 52.36 & 18.52 & 70.27
& 70.72 & 52.80 & \textbf{66.72} \\
Qwen3-8B
& 66.00 & 24.60 & 0.00
& 24.60 & 8.60 & 0.00
& 0.65 & 2.15
& 22.40 & 10.46 & 13.27
& 14.87 & 19.20 & \textbf{15.91} \\
\bottomrule
\end{tabular}
\captionof{table}{Complete OjaKV-PF results (\%) on RULER at 16K and
$\rho=0.60$. Each task contains 500 examples.}
\label{tab:ojakv-ruler-16k}
\end{minipage}
\end{center}

\begin{center}
\begin{minipage}{\textwidth}
Table~\ref{tab:ojakv-longbench} reports the complete 21-task LongBench
results.
\medskip

\centering
\footnotesize
\renewcommand{\arraystretch}{1.12}
\textbf{(a) English-language and code tasks}
\par\smallskip
\setlength{\tabcolsep}{1.10pt}
\begin{tabular*}{\textwidth}{@{\extracolsep{\fill}}l*{17}{c}@{}}
\toprule
\textbf{Model}
& \multicolumn{3}{c}{\textbf{Single-Doc QA}}
& \multicolumn{3}{c}{\textbf{Multi-Doc QA}}
& \multicolumn{3}{c}{\textbf{Summarization}}
& \multicolumn{3}{c}{\textbf{Few-shot}}
& \multicolumn{2}{c}{\textbf{Synthetic}}
& \multicolumn{2}{c}{\textbf{Code}}
& \textbf{Avg.-16} \\
\cmidrule(lr){2-4}
\cmidrule(lr){5-7}
\cmidrule(lr){8-10}
\cmidrule(lr){11-13}
\cmidrule(lr){14-15}
\cmidrule(lr){16-17}
& \textbf{NQA} & \textbf{Qasp} & \textbf{MF-en}
& \textbf{HQA} & \textbf{2WQA} & \textbf{Mus}
& \textbf{GRep} & \textbf{QMS} & \textbf{MNews}
& \textbf{TREC} & \textbf{TQA} & \textbf{SAM}
& \textbf{PC} & \textbf{PR-en}
& \textbf{LCC} & \textbf{RB-P} & \\
\midrule
Llama-3.1-8B-Instruct
& 30.08 & 41.33 & 55.61
& 56.98 & 47.43 & 33.07
& 23.11 & 23.76 & 23.80
& 71.00 & 90.50 & 37.02
& 10.00 & 99.50
& 22.86 & 19.89 & \textbf{42.87} \\
Qwen3-8B
& 4.81 & 7.43 & 13.87
& 8.80 & 8.00 & 4.11
& 6.58 & 10.03 & 7.23
& 59.00 & 79.54 & 25.35
& 2.89 & 47.43
& 8.50 & 7.72 & \textbf{18.83} \\
\bottomrule
\end{tabular*}

\medskip
\textbf{(b) Chinese tasks}
\par\smallskip
\begin{tabular*}{0.78\textwidth}{@{\extracolsep{\fill}}lccccc cc@{}}
\toprule
\textbf{Model}
& \textbf{MF-zh} & \textbf{DuR} & \textbf{VCSum}
& \textbf{LSHT} & \textbf{PR-zh}
& \textbf{Avg.-16} & \textbf{Avg.-21} \\
\midrule
Llama-3.1-8B-Instruct
& 54.92 & 27.70 & 13.26 & 39.75 & 80.70
& 42.87 & \textbf{42.97} \\
Qwen3-8B
& 10.59 & 8.43 & 6.48 & 39.42 & 46.10
& 18.83 & \textbf{19.63} \\
\bottomrule
\end{tabular*}
\captionof{table}{Complete OjaKV-PF results (\%) on LongBench at $\rho=0.60$.
Panel (a) reports the 14 English-language and two code tasks; panel (b)
reports the five Chinese tasks. Avg.-16 and Avg.-21 average over the
corresponding task sets.}
\label{tab:ojakv-longbench}
\end{minipage}
\end{center}

Overall, the reproduced OjaKV-PF performance varies substantially across
models and tasks, with particularly large degradation and several zero
scores on Qwen3-8B.

\section{Design Ablations}
\label{app:additional-ablations}

\subsection{Page-Size Selection}

To isolate the effect of page size, we choose a symmetric rank
$r_K=r_V$ for each page size so that the factorized-page storage ratios
are closely matched. For a key--value page pair with head dimension
$d=128$, the retained ratio is
\begin{equation}
\rho_{\mathrm{factor}}(P,r_K,r_V)
=\frac{(r_K+r_V)(P+d)}{2Pd}.
\label{eq:page-size-factor-storage}
\end{equation}
The three configurations are therefore $(P,r_K,r_V)=(16,9,9)$,
$(32,16,16)$, and $(64,27,27)$. These ratios apply to factorized pages,
and all configurations use the same dense-page policy.
Table~\ref{tab:page-size-selection} reports all 13 RULER task scores
and decomposition time.

\begin{center}
\begin{minipage}{\textwidth}
\centering
\footnotesize
\renewcommand{\arraystretch}{1.16}
\setlength{\tabcolsep}{1.0pt}
\begin{tabular*}{\textwidth}{@{\extracolsep{\fill}}c*{15}{c}@{}}
\toprule
\multirow{2}{*}{\textbf{$(P,r_K,r_V)$}}
& \multicolumn{8}{c}{\textbf{Retrieval}}
& \multicolumn{1}{c}{\textbf{Multi-hop}}
& \multicolumn{2}{c}{\textbf{Agg.}}
& \multicolumn{2}{c}{\textbf{QA}}
& \multirow{2}{*}{\textbf{Avg.}}
& \textbf{Decomp. Time} \\
\cmidrule(lr){2-9}
\cmidrule(lr){10-10}
\cmidrule(lr){11-12}
\cmidrule(lr){13-14}
& \textbf{S1} & \textbf{S2} & \textbf{S3}
& \textbf{MK1} & \textbf{MK2} & \textbf{MK3}
& \textbf{MQ} & \textbf{MV} & \textbf{VT}
& \textbf{CWE} & \textbf{FWE}
& \textbf{QA-1} & \textbf{QA-2} & & \\
\midrule
$(16,9,9)$
& 99.20 & 97.20 & 79.60 & 97.00 & 99.60 & 88.00
& 97.05 & 90.25 & 93.84 & \textbf{85.86} & 88.73
& 70.17 & 55.00 & 87.81 & 0.11 s \\
$(32,16,16)^\dagger$
& 99.40 & 97.20 & 91.20 & 99.00 & \textbf{100.00} & 96.40
& \textbf{99.60} & 92.15 & 99.16 & \textbf{85.86} & 88.47
& 71.10 & 54.00 & 90.27 & 0.17 s \\
$(64,27,27)$
& \textbf{100.00} & \textbf{99.80} & \textbf{99.20}
& \textbf{99.60} & \textbf{100.00} & \textbf{98.80}
& \textbf{99.60} & \textbf{94.25} & \textbf{99.36}
& 85.16 & \textbf{88.80}
& \textbf{71.90} & \textbf{56.20} & \textbf{91.74} & 0.27 s \\
\bottomrule
\end{tabular*}
\captionof{table}{Page-size selection on Llama-3.1-8B-Instruct at 16K.
For each page size, we set $r_K=r_V$ and select the shared rank to keep
factorized-page storage near 63\%.
Decomposition time is the mean per sample during prefill; $\dagger$ marks
the main configuration.}
\label{tab:page-size-selection}
\end{minipage}
\end{center}

\subsection{Rank Selection}

Table~\ref{tab:rank-selection-full} reports the full rank grid while
holding page size and the dense-page policy fixed. Configurations are
grouped by total rank $r_K+r_V$.

\begingroup
\footnotesize
\renewcommand{\arraystretch}{1.16}
\setlength{\tabcolsep}{1.2pt}
\setlength{\LTleft}{0pt}
\setlength{\LTright}{0pt}
\begin{longtable}{@{\extracolsep{\fill}}cc*{14}{c}@{}}
\toprule
\multicolumn{2}{c}{\textbf{Ranks}}
& \multicolumn{8}{c}{\textbf{Retrieval}}
& \multicolumn{1}{c}{\textbf{Multi-hop}}
& \multicolumn{2}{c}{\textbf{Agg.}}
& \multicolumn{2}{c}{\textbf{QA}}
& \multicolumn{1}{c}{} \\
\cmidrule(lr){1-2}
\cmidrule(lr){3-10}
\cmidrule(lr){11-11}
\cmidrule(lr){12-13}
\cmidrule(lr){14-15}
\textbf{$r_K$} & \textbf{$r_V$} &
\textbf{S1} & \textbf{S2} &
\textbf{S3} & \textbf{MK1} & \textbf{MK2} & \textbf{MK3} &
\textbf{MQ} & \textbf{MV} & \textbf{VT} & \textbf{CWE} &
\textbf{FWE} & \textbf{QA-1} & \textbf{QA-2} & \textbf{Avg.} \\
\midrule
\endfirsthead
\toprule
\multicolumn{2}{c}{\textbf{Ranks}}
& \multicolumn{8}{c}{\textbf{Retrieval}}
& \multicolumn{1}{c}{\textbf{Multi-hop}}
& \multicolumn{2}{c}{\textbf{Agg.}}
& \multicolumn{2}{c}{\textbf{QA}}
& \multicolumn{1}{c}{} \\
\cmidrule(lr){1-2}
\cmidrule(lr){3-10}
\cmidrule(lr){11-11}
\cmidrule(lr){12-13}
\cmidrule(lr){14-15}
\textbf{$r_K$} & \textbf{$r_V$} &
\textbf{S1} & \textbf{S2} &
\textbf{S3} & \textbf{MK1} & \textbf{MK2} & \textbf{MK3} &
\textbf{MQ} & \textbf{MV} & \textbf{VT} & \textbf{CWE} &
\textbf{FWE} & \textbf{QA-1} & \textbf{QA-2} & \textbf{Avg.} \\
\midrule
\endhead
\midrule
\multicolumn{16}{r}{\textit{Continued on next page}} \\
\endfoot
\bottomrule
\endlastfoot
12 & 12 & 74.80 & 68.40 & 15.60 & 72.60 & 83.80 & 46.80 & 71.30 & 54.05 & 71.40 & 67.32 & 89.00 & 67.43 & 53.20 & 64.28 \\
\noalign{\vskip 1.2pt}\hdashline\noalign{\vskip 1.2pt}
12 & 14 & 80.60 & 72.20 & 28.60 & 79.40 & 86.00 & 63.80 & 77.50 & 60.75 & 76.84 & 69.88 & 89.27 & 69.03 & 53.20 & 69.77 \\
14 & 12 & 91.60 & 86.20 & 37.60 & 90.20 & 97.20 & 75.60 & 91.25 & 77.65 & 91.76 & 80.24 & 89.60 & 69.13 & 52.80 & 79.29 \\
\noalign{\vskip 1.2pt}\hdashline\noalign{\vskip 1.2pt}
12 & 16 & 83.80 & 76.60 & 40.60 & 80.00 & 88.00 & 70.80 & 81.00 & 65.05 & 82.20 & 71.18 & 89.13 & 69.53 & 53.40 & 73.18 \\
14 & 14 & 94.40 & 89.60 & 61.40 & 92.80 & 99.00 & 85.40 & 93.85 & 82.35 & 95.64 & 82.08 & 89.07 & 70.80 & 53.40 & 83.83 \\
16 & 12 & 98.80 & 95.60 & 60.00 & 97.80 & 100.00 & 86.80 & 98.45 & 89.60 & 97.80 & 83.20 & 88.53 & 70.43 & 53.20 & 86.17 \\
\noalign{\vskip 1.2pt}\hdashline\noalign{\vskip 1.2pt}
12 & 18 & 85.60 & 78.00 & 49.60 & 83.60 & 89.20 & 74.20 & 83.25 & 68.20 & 83.52 & 70.82 & 89.20 & 69.43 & 53.40 & 75.23 \\
14 & 16 & 96.20 & 91.40 & 73.80 & 95.20 & 98.60 & 89.20 & 94.85 & 85.15 & 96.76 & 82.66 & 89.20 & 71.07 & 53.20 & 85.95 \\
\textbf{16} & \textbf{14$^{\dagger}$} & 98.80 & 95.80 & 82.00 & 98.20 & 100.00 & 93.40 & 98.85 & 91.10 & 98.80 & 84.84 & 88.40 & 70.83 & 53.60 & \textbf{88.82} \\
18 & 12 & 99.80 & 98.60 & 75.60 & 99.60 & 100.00 & 91.00 & 99.80 & 94.95 & 99.04 & 83.92 & 88.27 & 70.60 & 53.60 & 88.83 \\
\noalign{\vskip 1.2pt}\hdashline\noalign{\vskip 1.2pt}
12 & 20 & 87.20 & 80.80 & 58.00 & 84.60 & 88.40 & 75.40 & 83.70 & 70.35 & 85.16 & 72.64 & 89.33 & 69.27 & 53.60 & 76.80 \\
14 & 18 & 97.00 & 92.80 & 82.20 & 96.40 & 98.80 & 91.80 & 96.00 & 86.55 & 96.48 & 83.20 & 88.93 & 71.07 & 54.00 & 87.33 \\
16 & 16 & 99.40 & 97.20 & 91.20 & 99.00 & 100.00 & 96.40 & 99.60 & 92.15 & 99.16 & 85.86 & 88.47 & 71.10 & 54.00 & 90.27 \\
18 & 14 & 100.00 & 99.20 & 90.40 & 99.60 & 100.00 & 96.00 & 99.50 & 95.65 & 99.32 & 85.36 & 88.33 & 71.27 & 54.20 & 90.68 \\
20 & 12 & 100.00 & 100.00 & 78.80 & 99.60 & 100.00 & 92.80 & 99.70 & 96.10 & 99.48 & 84.24 & 88.60 & 71.10 & 54.00 & 89.57 \\
\noalign{\vskip 1.2pt}\hdashline\noalign{\vskip 1.2pt}
14 & 20 & 98.00 & 93.00 & 85.00 & 96.60 & 98.80 & 93.20 & 96.45 & 87.35 & 97.08 & 84.28 & 89.20 & 71.30 & 54.00 & 88.02 \\
16 & 18 & 99.60 & 98.00 & 95.00 & 99.20 & 100.00 & 97.60 & 99.55 & 93.15 & 99.28 & 86.44 & 88.33 & 71.50 & 53.80 & 90.88 \\
18 & 16 & 100.00 & 99.60 & 96.20 & 99.60 & 100.00 & 98.20 & 99.70 & 94.80 & 99.40 & 86.78 & 88.27 & 71.67 & 54.40 & 91.43 \\
20 & 14 & 100.00 & 100.00 & 93.60 & 99.60 & 100.00 & 97.20 & 99.80 & 97.40 & 99.68 & 86.34 & 88.33 & 71.10 & 54.20 & 91.33 \\
\noalign{\vskip 1.2pt}\hdashline\noalign{\vskip 1.2pt}
16 & 20 & 99.80 & 98.60 & 96.00 & 99.60 & 100.00 & 96.80 & 99.85 & 93.35 & 99.52 & 86.74 & 88.53 & 72.23 & 53.80 & 91.14 \\
18 & 18 & 100.00 & 99.80 & 98.00 & 99.60 & 100.00 & 98.40 & 99.85 & 96.35 & 99.56 & 87.18 & 88.33 & 72.17 & 54.00 & 91.79 \\
20 & 16 & 100.00 & 100.00 & 97.00 & 99.60 & 100.00 & 98.40 & 99.80 & 97.70 & 99.68 & 86.46 & 88.47 & 71.63 & 54.40 & 91.78 \\
\noalign{\vskip 1.2pt}\hdashline\noalign{\vskip 1.2pt}
18 & 20 & 100.00 & 99.80 & 99.40 & 99.60 & 100.00 & 98.40 & 99.70 & 95.90 & 99.52 & 87.70 & 88.53 & 72.30 & 54.60 & 91.96 \\
20 & 18 & 100.00 & 100.00 & 99.20 & 99.60 & 100.00 & 98.40 & 99.65 & 97.60 & 99.68 & 86.58 & 89.00 & 71.70 & 54.00 & 91.95 \\
\noalign{\vskip 1.2pt}\hdashline\noalign{\vskip 1.2pt}
20 & 20 & 100.00 & 100.00 & 99.80 & 99.60 & 100.00 & 98.60 & 99.50 & 97.55 & 99.72 & 87.52 & 88.60 & 71.97 & 54.60 & 92.11 \\
\end{longtable}
\addtocounter{table}{-1}
\captionof{table}{Complete RULER rank-selection results (\%) for
Llama-3.1-8B-Instruct at 16K. Every configuration uses $P=32$ and the
same dense-page policy. Dashed rules group equal total ranks
$r_K+r_V$; $\dagger$ marks the main configuration.}
\label{tab:rank-selection-full}
\endgroup

\subsection{RoPE Key Analysis}
\label{app:pre-rope}

Table~\ref{tab:rope-selection} compares pre- and post-RoPE keys at 16K,
$P=32$, and $(r_K,r_V)=(16,16)$. Pre-RoPE raises the average score from
90.40 to 90.88.

\begin{center}
\begin{minipage}{\textwidth}
\centering
\footnotesize
\setlength{\tabcolsep}{5pt}
\begin{tabular*}{\textwidth}{@{\extracolsep{\fill}}l*{14}{c}@{}}
\toprule
\multicolumn{1}{c}{}
& \multicolumn{8}{c}{\textbf{Retrieval}}
& \multicolumn{1}{c}{\textbf{Multi-hop}}
& \multicolumn{2}{c}{\textbf{Agg.}}
& \multicolumn{2}{c}{\textbf{QA}}
& \multicolumn{1}{c}{} \\
\cmidrule(lr){2-9}
\cmidrule(lr){10-10}
\cmidrule(lr){11-12}
\cmidrule(lr){13-14}
\textbf{Key} & \textbf{S1} & \textbf{S2} & \textbf{S3}
& \textbf{MK1} & \textbf{MK2} & \textbf{MK3}
& \textbf{MQ} & \textbf{MV} & \textbf{VT}
& \textbf{CWE} & \textbf{FWE}
& \textbf{QA-1} & \textbf{QA-2} & \textbf{Avg.} \\
\midrule
\textbf{Post-RoPE}
& 99.40 & 97.20 & 92.00 & 99.00 & 99.80 & 96.60
& 99.75 & 93.05 & 99.24 & 84.80 & 88.53
& 71.23 & \textbf{54.60} & 90.40 \\
\textbf{Pre-RoPE}
& \textbf{100.00} & \textbf{99.00} & \textbf{93.80}
& 99.00 & \textbf{100.00} & \textbf{97.20}
& 99.75 & \textbf{93.10} & \textbf{99.28}
& \textbf{86.02} & \textbf{88.93}
& \textbf{71.33} & 54.00 & \textbf{90.88} \\
\bottomrule
\end{tabular*}
\captionof{table}{Pre- versus post-RoPE key factorization on
Llama-3.1-8B-Instruct at 16K. Scores are percentages over the 13 RULER
tasks; QA-1 and QA-2 denote SQuAD and HotpotQA.}
\label{tab:rope-selection}
\end{minipage}
\end{center}